\documentclass[letterpaper]{article} 
\usepackage{aaai2027}  

\nocopyright

\usepackage[hyphens]{url}  
\usepackage{graphicx} 
\usepackage{natbib}  
\usepackage{caption} 
\usepackage{algorithm}
\usepackage{algorithmic}

\usepackage{newfloat}
\usepackage{listings}
\DeclareCaptionStyle{ruled}{labelfont=normalfont,labelsep=colon,strut=off} 
\floatstyle{ruled}
\newfloat{listing}{tb}{lst}{}
\floatname{listing}{Listing}

\usepackage{booktabs}

\usepackage{amsmath, amsfonts, amssymb}

\usepackage{tikz}
\usetikzlibrary{arrows.meta,decorations.pathreplacing,positioning,calc,fit}
\usepackage{subcaption}

		\title{A Semi-Analytical Inference Framework for Structural Causal Models}
		
		\author {
			Tatiana Boura\corresponding,
			Robert Peharz
		}
		\affiliations {
			Graz University of Technology
		}

\begin{document}
			
			\maketitle
			
			\begin{abstract}
				Structural Causal Models (SCMs) provide a principled framework for observational, interventional, and counterfactual reasoning, all of which rely on probabilistic inference. Existing frameworks either transform SCMs into alternative tractable representations or rely on approximate inference with expressive generative models. We instead perform inference using a message passing scheme directly on the SCM by representing structural mechanisms and probability densities in a common finite-dimensional function space. Forward propagation combines a numerical realization of the Law of the Unconscious Statistician, while propagated densities are recovered by spline projection from LOTUS moments. The same representation naturally yields backward message passing for conditioning, providing a unified framework for causal inference. Using tensor-product B-spline bases, the resulting algorithm expresses inference through numerical quadrature and linear algebra without explicitly constructing conditional distributions. Experiments on randomly generated continuous SCMs demonstrate accurate approximation of Monte Carlo reference solutions and systematic improvement with increasing basis resolution.
			\end{abstract}
			
			
			\section{Introduction}
			
			The notion of causality has been central to philosophical discourse since antiquity.  Structural Causal Models (SCMs) provide one of the most widely used mathematical frameworks for causal reasoning~\cite{pearl:2009}. Under the assumption that the modeled system is causally closed, uncertainty is confined to exogenous variables and endogenous variables are related deterministically~\cite{peters:2017}.
			SCMs support three levels of reasoning, known as Pearl's causal hierarchy: \emph{association}, \emph{intervention}, and \emph{counterfactuals}~\cite{book_of_why:2018}. Although these three query types differ in their semantics, each ultimately requires evaluating probability distributions induced by the SCM. Efficient probabilistic inference therefore forms the computational core of causal reasoning. Observational queries require probabilistic inference in the observational distribution, i.e.~the pushforward distribution of the exogenous distribution under the hierarchical causal mechanism,
			interventional queries require inference after modifying the structural equations, and counterfactual reasoning combines conditioning with hypothetical interventions~\cite{complexity_counterfactuals:2023,pearl:2009}.
			Specifically, counterfactual inference can be formulated using a twin SCM or, equivalently, an abduct-intervene-infer procedure~\cite{pearl:2009}. 
			
			While much of the causal modeling literature focuses on causal discovery, identification, and effect estimation from data~\cite{peters:2017}, comparatively less attention has been devoted to efficiently evaluating queries in a given SCM~\cite{causal_circuits:2022,wang:2023,ispn:2021, pawlowski:2020}. These directions are complementary: discovery and identification determine the causal model and identifiable effects, whereas causal inference assumes a given SCM and answers causal queries. This setting naturally arises when the model is provided by expert knowledge or a preceding learning stage.
			
			Causal inference appears closely related to inference in Bayesian networks (BNs)~\cite{pgm_book:2009}, where algorithms such as variable elimination, belief propagation, and junction-tree message passing enable efficient exact and approximate inference~\cite{pearl:1982,ve:1996,wainwright:2008}. Since every acyclic SCM induces a BN, it is natural to ask whether these methods apply directly. However, BN inference requires explicit conditional distributions, whereas SCMs define them only implicitly through deterministic structural mechanisms and exogenous noise. Thus, inference depends both on the graph structure and the representation of the structural mechanisms.
			
			Several approaches have addressed this challenge. One line of work compiles SCMs into alternative representations that support tractable inference, including tractable circuits and related representations~\cite{causal_circuits:2022,wang:2023,ispn:2021}. Another employs expressive structural mechanisms based on deep generative models~\cite{pawlowski:2020}, where inference is typically performed approximately. Existing methods therefore either transform the SCM into a tractable representation or perform inference approximately using sampling or learned generative models.
			
			In this work, we instead perform inference directly on the original SCM representation. We formulate message passing in a common finite-dimensional function space for structural mechanisms and probability densities. Forward propagation evaluates the Law of the Unconscious Statistician (LOTUS) 
			integrals numerically and recovers propagated densities through spline projection, yielding a semi-analytical inference procedure that avoids SCM compilation and sampling-based inference. The same representation naturally extends to backward message passing, providing a unified framework for observational, interventional, and counterfactual inference.
			
			We instantiate the framework using tensor-product B-spline bases, whose approximation properties, numerical stability, and efficient quadrature algorithms~\cite{de_boor:2001,schumaker:2007,isogeometric_analysis:2009} allow inference to be expressed through linear algebra and numerical quadrature while representing highly nonlinear mechanisms.
			Specifically, we model both densities and structural mechanisms using tensor-product B-splines. 
			Restricting structural mechanisms is not itself new~\cite{hoyer:2008,buhlmann:2014,morville:2025,zhang:2009}. Previous work, however, does so primarily to facilitate causal discovery or parameter estimation. In contrast, we employ a restricted mechanism family to perform efficient inference in fully specified continuous SCMs.
			The main contributions of this work are as follows:
			\begin{itemize}
				\item We formulate probabilistic inference in continuous SCMs as message passing in a common finite-dimensional approximate functional space, avoiding the explicit construction of the  distributions induced by nonlinear structural mechanisms.
				\item We derive forward and backward message-passing operators based on LOTUS and spline projection, reducing inference to linear algebra and numerical quadrature in this common function space.
				\item We instantiate the framework using tensor-product B-spline representations for both structural mechanisms and probability densities, resulting in a practical semi-analytical inference algorithm for continuous SCMs.
			\end{itemize}
			
			\section{Preliminaries}
			
			We briefly review the concepts required throughout the paper and establish the notation.
			
			\subsection{Structural Causal Models}\label{sec:scms}
			
			A \emph{Structural Causal Model} (SCM)~\cite{pearl:2009} is a tuple
			\[
			\mathcal{M}=(\mathcal{U},\mathcal{V},\mathcal{F},P(\mathcal{U})),
			\]
			where
			\begin{itemize}
				\item $\mathcal{U}$ is the set of exogenous variables;
				\item $\mathcal{V}$ is the set of endogenous variables;
				\item $\mathcal{F}=\{f_X\}_{X\in\mathcal{V}}$ is a collection of structural mechanisms,
				\[
				X=f_X(\operatorname{Pa}(X),U_X),
				\]
				where each endogenous variable is a deterministic function of its endogenous parents, \(\operatorname{Pa}(X)\), and its associated exogenous variable \(U_X\). Throughout this work, we assume mutually independent exogenous variables and one exogenous variable per endogenous variable.
				\item $P(\mathcal{U})$ is the joint probability distribution of the exogenous variables.
			\end{itemize}
			
			The structural mechanisms induce a directed acyclic graph (DAG) $\mathcal{G}$ whose vertices are the
			endogenous variables and whose edges encode parent-child relationships. Under
			the above assumptions, the SCM satisfies the causal Markov
			property~\cite{peters:2017}, implying the factorization
			\(
			P(\mathcal{V})
			=
			\prod\nolimits_{X\in\mathcal{V}}
			P(X\mid\operatorname{Pa}(X)).
			\)
			
			SCMs support three principal classes of inference queries~\cite{book_of_why:2018}.
			\emph{Observational} queries compute probabilities conditioned on observed
			variables. \emph{Interventional} queries evaluate the distribution resulting
			from modifying one or more structural mechanisms via the \(do(\cdot)\)
			operator. \emph{Counterfactual} queries reason about hypothetical outcomes by
			combining observed evidence with interventions on the same SCM instance. Although
			these queries differ in their semantics, they all ultimately require
			efficient probabilistic inference. 
			For an illustrative representation of these queries, we refer the reader to the Appendix.

			\subsection{Message Passing on Graphical Models}\label{sec:message_passing}
			
			Exact probabilistic inference in graphical models is commonly performed by
			message-passing algorithms, which exploit conditional independence to
			decompose global inference into local computations
			\cite{pn_book:1999,pgm_book:2009}. Rather than manipulating the full joint
			distribution directly, neighboring regions exchange messages that summarize the
			influence of the remaining variables.
			
			For tree-structured graphical models, exact marginals are computed by the
			sum-product algorithm, which propagates messages along the graph
			edges~\cite{pearl:1982}. For non-tree models, exact inference first
			constructs a junction tree whose nodes correspond to variable cliques. Messages
			are then exchanged between neighboring cliques through their separator sets
			using local multiplication, marginalization, and normalization~\cite{junction_tree:1988,jensen:1990,shenoy_shafer:1990}.
			
			Let
			\(
			\mathcal T=(\mathcal C,\mathcal E)
			\)
			denote a junction tree, where each node
			\(C\in\mathcal C\) is a clique and each edge
			\((C_i,C_j)\in\mathcal E\) connects neighboring cliques. The separator
			associated with an edge is
			\(
			S_{ij}=C_i\cap C_j.
			\)
			Each clique stores a joint state, while separator messages ensure consistency between neighboring cliques. 
			After the collect and distribute passes, calibrated clique states agree on their separator marginals.
			
			Although every acyclic SCM induces a Bayesian network~\cite{pgm_book:2009}
			satisfying the causal Markov property (Section~\ref{sec:scms}), its local
			conditional distributions are not represented explicitly. Instead, they are
			defined implicitly by deterministic structural mechanisms and exogenous
			variables. Consequently, the local probability factors required by standard
			message-passing algorithms are unavailable.
			The proposed framework overcomes this limitation by propagating probability
			measures directly through structural mechanisms rather than explicit
			conditional densities. By combining numerical realizations of the LOTUS with
			least-squares recovery in a common functional space, the resulting
			algorithm retains the junction-tree message-passing paradigm while avoiding the
			explicit construction of conditional probability distributions.

			\subsection{Spline Representation}\label{sec:splines}
			
			Throughout, probability densities and structural mechanisms are
			represented in a finite-dimensional B-spline space. Let
			\[
			\{N_{i}\}_{i=1}^{n}
			\]
			denote a B-spline basis of polynomial degree \(3\) defined by the Cox-de Boor formula~\cite{de_boor:2001} over the
			computational domain \(\Omega\). Any spline in this space can be written as
			\[
			s(u)
			=
			\sum\nolimits_{i=1}^{n}
			w_iN_{i}(u),
			\]
			where $w_i\in\mathbb{R}$ are the spline coefficients. 
			
			The framework relies on two fundamental properties of B-splines. First,
			\(N_{i}(x)\ge0\), so non-negative coefficients produce non-negative
			functions. Second, the basis functions are piecewise polynomials, enabling
			efficient evaluation of integrals by Gaussian quadrature.
			
			Let $f \colon \Omega\rightarrow\mathbb{R}$ be approximated in the B-spline space
			spanned by \(\{N_i\}_{i=1}^{n}\). The approximation is
			\[
			\hat f(x)
			=
			\sum\nolimits_{i=1}^{n}
			w_iN_i(x)
			=
			\mathbf N(x)^\top\mathbf w.
			\]
			The spline coefficients are obtained by the $L^2(\Omega)$ projection~\cite{isogeometric_analysis:2009}, which yields the linear system
			\[
			\mathbf M\mathbf w=\mathbf b,
			\]
			where
			\(
			M_{ij}
			=
			\int_\Omega
			N_i(\mathbf{x})
			N_j(\mathbf{x})
			\,d\mathbf{x}
			\)
			form the \emph{mass matrix} and
			\(
			b_i
			=
			\int_\Omega
			N_i(\mathbf{x})f(\mathbf{x})
			\,d\mathbf{x}
			\)
			the \emph{load vector}.
			Since the B-spline basis is linearly independent, the mass matrix is symmetric
			positive definite and the projection is uniquely defined. Both the mass matrix
			and load vector are evaluated efficiently using Gaussian quadrature. A
			\(q\)-point Gaussian quadrature rule integrates polynomials of degree up to
			\(2q-1\) exactly~\cite{quad_exactness:2007}.
			More details in the Appendix.
			
			Multivariate spline spaces are obtained by tensor products,
			\[
			N_{\mathbf{i}}(\mathbf{x})
			=
			\prod\nolimits_{k=1}^{d}
			N^{(k)}_{i_k}(x_k),
			\]
			where $\mathbf{i}=(i_1,\ldots,i_d)$ is a multi-index, giving
			\[
			s(\mathbf{x})
			=
			\sum\nolimits_{i_1=1}^{n_1}
			\cdots
			\sum\nolimits_{i_d=1}^{n_d}
			w_{i_1,\ldots,i_d}
			\prod\nolimits_{k=1}^{d}
			N^{(k)}_{i_k}(x_k),
			\]
			where $w_{i_1,\ldots,i_d}$ are the spline coefficients. The resulting basis inherits the properties of one-dimensional B-splines in each coordinate direction~\cite{nurbs:1997,isogeometric_the_begining:2005}.
			The tensor-product structure implies
			factorization of the  multidimensional mass matrix into products of one-dimensional matrices, allowing both the
			projection and Gaussian quadrature to be performed efficiently by tensor
			products of one-dimensional rules~\cite{quadrature:2002}. 
			These projection operators form the numerical foundation of this inference framework.

			\section{Method}
			
			This section presents the message-passing framework for inference
			in fully specified SCMs. Whereas causal discovery and identification determine the causal model
			and identifiable causal effects, we address the complementary problem of efficiently
			answering causal queries. The algorithm operates directly on structural
			mechanisms, avoiding the explicit construction of conditional densities.
			Inference proceeds in two stages. The forward pass propagates probability
			measures through the SCM to construct marginals and clique states for
			observational and interventional inference. The backward pass then propagates
			likelihood information through these states to incorporate evidence, enabling
			conditional and counterfactual inference.
			
			\subsection{Densities and Mechanisms as Splines}\label{sec:spline_representation}
			
			The framework represents probability densities and structural mechanisms in a common finite-dimensional spline space, 
			allowing all inference operations to be expressed through spline coefficients, projection operators, and numerical quadrature.
			Let $\operatorname{Pa}(X)$
			denote the parent set of an endogenous variable \(X\) and \( |\operatorname{Pa}(X) | = M\). For each parent variable \(X_m  \in \operatorname{Pa}(X)\), let
			\(
			\{N^{(m)}_{i}\}_{i=1}^{n_m}
			\)
			be a one-dimensional B-spline basis defined over the computational domain
			\(\Omega_m=[\alpha_m,\beta_m]\). The tensor-product basis associated with the parent set is 
			\[
			N_{\mathbf{j}}(\mathbf{x})
			=
			\prod\nolimits_{m=1}^{M}
			N^{(m)}_{j_m}(x_m),
			\]
			where
			\(\mathbf{x}=(x_1,\ldots,x_M)\)
			and
			\(\mathbf{j}=(j_1,\ldots,j_M)\).
			The joint marginal density over the parent variables is represented by
			\begin{align*}
				p_{\operatorname{Pa}(X)}(\mathbf{x})
				&=
				\sum\nolimits_{\mathbf{j}}
				w^{\operatorname{Pa}(X)}_{\mathbf{j}}
				N_{\mathbf{j}}(\mathbf{x})\\
				&=
				\sum_{j_1=1}^{n_1}
				\cdots
				\sum_{j_M=1}^{n_M}
				w^{\operatorname{Pa}(X)}_{j_1,\ldots,j_M}
				\prod_{m=1}^{M}
				N^{(m)}_{j_m}(x_m).
			\end{align*}
			The coefficients are collected in the tensor
			\(\mathbf w_{\operatorname{Pa}(X)}\), which characterizes the modeled
			joint density~\cite{densities_as_splines:2012}.
			
			A tensor-product B-spline expansion is generally not a probability density.
			Even so, non-negative coefficients guarantee a
			non-negative spline (Section~\ref{sec:splines}). Its total mass is
			\[
			\int_{\Omega_{\operatorname{Pa}(X)}}\!\!\!\!\!\!\!\!\!
			p_{\operatorname{Pa}(X)}(\mathbf{x})\,d\mathbf{x}
			=
			\sum\nolimits_{\mathbf{j}}
			w_{\mathbf{j}}^{\operatorname{Pa}(X)} c_{\mathbf{j}},
			\ \
			c_{\mathbf{j}}
			=
			\int_{\Omega_{\operatorname{Pa}(X)}}\!\!\!\!\!\!\!\!\!
			N_{\mathbf{j}}(\mathbf{x})\,d\mathbf{x},
			\]
			where \(c_{\mathbf j}\) are the tensor-product mass coefficients. Hence, the
			spline defines a valid probability density whenever
			\[
			w_{\mathbf{j}}^{\operatorname{Pa}(X)}\ge0,
			\qquad
			\sum\nolimits_{\mathbf{j}}w_{\mathbf{j}}^{\operatorname{Pa}(X)}c_{\mathbf{j}}=1.
			\]
			
			The structural mechanism associated with an endogenous variable \(X\) is likewise represented using a tensor-product B-spline basis over its parent variables and the corresponding exogenous variable \(U_X\). Let
			\(
			\{N^{(U_X)}_{i}\}_{i=1}^{n_U}
			\)
			be a one-dimensional B-spline basis defined over the computational domain
			\(\Omega_{U_X}=[\alpha_U,\beta_U]\). 
			Then
			\begin{align*}
				f_X(\mathbf{x},u)
				={}&
				\sum_{\mathbf{r}}
				\sum_{\ell=1}^{n_U}
				\theta^X_{\mathbf{r},\ell}\,
				N_{\mathbf{r}}(\mathbf{x})
				N^{(U_X)}_{\ell}(u) \\
				={}&
				\sum_{r_1=1}^{n_1}
				\cdots
				\sum_{r_M=1}^{n_M}
				\sum_{\ell=1}^{n_U}
				\theta^X_{r_1,\ldots,r_M,\ell}
				\\
				&\times
				\left(
				\prod\nolimits_{m=1}^{M}
				N^{(m)}_{r_m}(x_m)
				\right)
				N^{(U_X)}_{\ell}(u).
			\end{align*}
			
			Throughout, we assume that the pushforward distributions
			considered by the method are absolutely continuous with respect to Lebesgue
			measure on their computational domains. 
			
			\subsection{Forward Message Passing}\label{sec:forward}
			The forward pass processes the SCM in topological order. For each endogenous 
			variable, it propagates the parent measure through the structural 
			mechanism to recover the child marginal. Whenever required to preserve downstream dependencies, 
			it additionally constructs joint spline states over selected variable scopes. 
			After forward propagation, scopes that are strict subsets of another 
			stored scope are discarded, and the remaining maximal scopes become 
			the cliques of the junction tree used during backward inference.
			
			Rather than explicitly constructing the pushforward density induced by a
			structural mechanism, the method characterizes the propagated measure
			through its moments. By the LOTUS, these moments can be evaluated for any
			measurable test function \(\varphi\)~\cite{pollard:2001}. Thus, for an
			endogenous variable \(X\) with structural mechanism \(f_X\),
			\begin{align}
				\mathbb{E}[\varphi(X)]
				=
				\mathbb{E}\!\left[
				\varphi\!\left(
				f_X(\operatorname{Pa}(X),U_X)
				\right)
				\right].
				\label{eq:exp}
			\end{align}
			It remains to choose test functions that recover the spline coefficients.
			Thus, the forward pass consists of three steps. 
			First, LOTUS is used to evaluate the moments of the propagated probability measure. 
			Second, these moments are projected onto the spline basis to recover a density. 
			Third, whenever dependencies must be preserved, the same procedure constructs
			joint clique states used later during backward inference.

			\paragraph{Marginal propagation.}
			The natural choice is to select the B-spline basis functions themselves as test functions,
			since the resulting moments are the entries of the load vector required by
			the \(L^2\) projection (Section~\ref{sec:splines}). Consequently, the
			forward pass computes the quantities needed to recover the spline
			coefficients.
			Given B-splines 
			\(\{N_i^X\}_{i=1}^{n_X}\) as test functions. Given the joint density \(p_{\operatorname{Pa}(X)}\) and \(f_X\colon  \Omega_{\operatorname{Pa(X)}} \! \times \Omega_{U_X} \! \! \rightarrow \Omega_X\), Eq.~(\ref{eq:exp}) reduces to
			\begin{align}
				V_i^X
				\!= \!\!
				\int_{\Omega_{\operatorname{Pa}(X)}}\!\!\!\!\!\!\!\!\!\!\!\!
				p_{\operatorname{Pa}(X)}(\mathbf x)
				\int_{\Omega_{U_X}}\!\!\!\!\!\!\!
				p_{U_X}(u)\,
				N_i^X\!\left(f_X(\mathbf x,u)\right)
				du\,d\mathbf x,
				\label{eq:forward_marginal_moments}
			\end{align}
			for \(i=1,\ldots,n_X\). 
			Let
			\(
			p_X(x)
			=
			\sum_{j=1}^{n_X}
			w_j^X N_j^X(x)
			\)
			be the spline approximation of the propagated density. Since
			\[
			\int_{\Omega_X}
			N_i^X(x)p_X(x)\,dx
			=
			\mathbb{E}\!\left[N_i^X(X)\right]
			=
			V_i^X,
			\]
			the projection coefficients satisfy
			\[
			\mathbf M_X\mathbf w_X=\mathbf V_X,
			\]
			where
			\(
			(\mathbf M_X)_{ij}
			=
			\int_{\Omega_X}
			N_i^X(x)N_j^X(x)\,dx
			\) are the mass matrix entries
			and
			\(\mathbf V_X = \left(V_1^X,\ldots,V_{n_X}^X\right)^\top\) is the load vector.

			\paragraph{Clique-state construction.}
			Marginal densities suffice only for single-variable queries. Message passing
			requires joint clique states over the junction tree introduced in
			Section~\ref{sec:message_passing}.  These joint
			states preserve the dependencies required by downstream structural mechanisms
			and later serve as the clique states updated during the backward pass.
			
			For an endogenous variable \(X\), let
			\(
			\mathbf R=\{R_m\}_{m=1}^{L}\subseteq\operatorname{Pa}(X)
			\)
			denote the parent variables retained in the joint state associated with
			\(X\), as determined by the required downstream dependencies. The resulting clique state has scope
			\(
			C= \mathbf R \cup \{X\}.
			\)
			Let
			\(
			\{N^{R_m}_{i_m}\}_{i_m=1}^{n_{R_m}}
			\)
			and
			\(
			\{N^X_i\}_{i=1}^{n_X}
			\)
			denote the corresponding one-dimensional B-spline bases.
			The tensor-product basis on the clique \(C\) is
			\[
			N_{\boldsymbol{i}}^C(\mathbf r,x)
			=
			\left(
			\prod\nolimits_{m=1}^{L}
			N_{i_m}^{R_m}(r_m)
			\right)
			N_{i_{L+1}}^X(x),
			\]
			where
			\(
			\boldsymbol{i}
			=
			(i_1,\ldots,i_L,i_X).
			\)
			Choosing the tensor-product basis functions as test
			functions yields the load-tensor entries
			\begin{align}
				V_{\boldsymbol{i}}^C
				&=
				\mathbb{E}\!\left[
				N_{\boldsymbol{i}}^C(\mathbf R,X)
				\right]
				\nonumber
				= \int_{\Omega_{\operatorname{Pa}(X)}}\!\!\!
				p_{\operatorname{Pa}(X)}(\mathbf x)
				\int_{\Omega_{U_X}}\!\!\!\!
				p_{U_X}(u) \\
				&\times
				\left(
				\prod\nolimits_{m=1}^{L}\!\!
				N_{i_m}^{R_m}(r_m)
				\right) 
				N_{i_{L+1}}^X
				\!\left(
				f_X(\mathbf x,u)
				\right)
				\,du\,d\mathbf x.
				\label{eq:forward_joint_moments}
			\end{align}
			The expectation is evaluated over the full parent tuple, while the clique state
			retains only the variables in \(C\). Variables in
			\(\operatorname{Pa}(X)\setminus\mathbf R\) are marginalized automatically, since
			their basis functions do not appear in the test function. Computationally, 
			this corresponds to quadrature summation over the omitted parent dimensions.
			The clique-state coefficients satisfy
			\(
			\mathbf M_C\mathbf w_C=\mathbf V_C,
			\)
			where,
			\(
			\mathbf M_C=
			\mathbf M_{R_1}\otimes\cdots\otimes
			\mathbf M_{R_L}\otimes
			\mathbf M_X.
			\)
			Its Kronecker structure enables matrix-free matrix-vector products, avoiding
			explicit construction of the full mass matrix.
			
			\paragraph{Clique-tree construction.}
			Since non-maximal scopes can be recovered by marginalization, only maximal
			scopes are retained. Let \(\mathcal S\) denote the collection of scopes for
			which the forward pass has constructed explicit joint spline states. Then
			\[
			\mathcal C
			=
			\left\{
			C\in\mathcal S:
			\nexists C'\in\mathcal S\text{ such that }C\subsetneq C'
			\right\}.
			\]
			The clique-intersection graph connects cliques with non-empty intersections,
			weighted by \(|S_{ij}|\). A maximum-weight spanning tree ~\cite{kruskal:1956,prim:1957} yields the junction
			tree~\cite{junction_tree:1988,jensen:1990}, which is accepted only if it
			satisfies the running-intersection property; that is, for every variable
			\(X\), the cliques containing \(X\) form a connected subtree.
			
			\paragraph{Quadrature and projection.}
			The moments in
			Eqs.~\eqref{eq:forward_marginal_moments}
			and~\eqref{eq:forward_joint_moments}
			are evaluated by tensor-product Gaussian quadrature. 
			The spline coefficients
			are then recovered by a regularized non-negative least-squares approximation (NNLS)~\cite{nnls:1995}
			of the corresponding \(L^2\)-projection system.
			Let
			\(
			\{x_{q_m}^{(m)},\omega_{q_m}^{(m)}\}_{q_m=1}^{Q_m}
			\),
			\(
			\{u_s,\omega_s^{(U)}\}_{s=1}^{S}
			\)
			denote the quadtrature rule for the parent and
			exogenous respectively. Defining
			\(
			\mathbf x_{\mathbf q}
			=
			\bigl(x_{q_1}^{(1)},\ldots,x_{q_M}^{(M)}\bigr)
			\)
			and
			\(
			\omega_{\mathbf q}
			=
			\prod\nolimits_{m=1}^{M}\omega_{q_m}^{(m)}
			\), 
			Eq.~\eqref{eq:forward_joint_moments} becomes
			\begin{align*}
				V_{\boldsymbol{i}}^C
				&\approx
				\sum\nolimits_{\mathbf q}
				\omega_{\mathbf q}
				p_{\operatorname{Pa}(X)}(\mathbf x_{\mathbf q})
				\sum\nolimits_{s=1}^{S}
				\omega_s^{(U)}
				p_{U_X}(u_s) \\
				&\times
				\left(
				\prod\nolimits_{m=1}^{L}
				N_{i_m}^{R_m}(r_{m,\mathbf q})
				\right)
				N_{i_{L+1}}^X
				\!\left(
				f_X(\mathbf x_{\mathbf q},u_s)
				\right).
			\end{align*}
			
			To enforce non-negative coefficients, we solve the diagonally regularized NNLS
			\[
			\mathbf w_C
			=
			\arg\min_{\mathbf w\geq0}
			\left\|
			(\mathbf M_C+\lambda\mathbf I)\mathbf w-\mathbf V_C
			\right\|_2^2,
			\]
			where the diagonal shift \(\lambda\mathbf I\) improves the numerical
			conditioning of the projection operator.
			The coefficients are then normalized,
			\(
			\mathbf w_C
			\leftarrow
			\frac{\mathbf w_C}
			{\sum_{\mathbf j} w_{\mathbf j} c_{\mathbf j}},
			\)
			ensuring that the recovered spline represents a normalized probability density.
			
			\paragraph{Interventions.}
			Interventional inference requires no additional message-passing operator. Under
			an intervention \(do(X=x^\star)\), the structural mechanism of \(X\) is
			replaced by the constant assignment \(X:=x^\star\), yielding a modified SCM.
			The forward pass is then reapplied, reusing the states of non-descendants and
			recomputing only those affected by the intervention. Since a point intervention
			induces a Dirac measure, the intervened variable is propagated
			deterministically rather than represented as a spline density.

			\subsection{Backward Message Passing}\label{sec:backward_message_passing}
			
			The backward pass incorporates observations by propagating likelihoods through
			the junction tree (Section~\ref{sec:message_passing}). Starting from the prior clique states 
			computed during the forward pass, neighboring cliques exchange likelihood-ratio messages 
			that update these priors into posterior clique states. Whenever a message is propagated through a structural
			mechanism, the corresponding conditional density is unavailable explicitly.
			Instead, likelihoods are propagated through the structural mechanism via a local
			backward operator. 
			
			\paragraph{Local backward operator.}
			
			Let \(X\) be an endogenous variable with structural mechanism \(f_X\) and 
			\(
			m_X:\Omega_X\rightarrow\mathbb{R}_{\ge0}
			\)
			denote the likelihood accumulated at \(X\).
			Propagating this likelihood through the structural mechanism induces the
			unnormalized likelihood
			\[
			\widetilde m_X(\mathbf{x},u)
			=
			m_X\!\left(f_X(\mathbf{x},u)\right),
			\]
			defined over the mechanism inputs.
			This representation can subsequently be marginalized onto any subset of the
			mechanism inputs.
			For ordinary backward message passing, the exogenous variable is marginalized,
			yielding the propagated likelihood over the endogenous parents,
			\[
			\mathcal{B}_X[m_X](\mathbf{x})
			=
			\int_{\Omega_{U_X}}
			\widetilde m_X(\mathbf{x},u)\,
			p_{U_X}(u)\,\mathrm du,
			\]
			where
			\(
			\mathbf{x}\in\Omega_{\operatorname{Pa}(X)}.
			\)
			Thus, evidence is propagated through the structural mechanism without
			explicitly constructing the conditional density
			\(
			p(X\mid\operatorname{Pa}(X)).
			\)
			Retaining the exogenous variable instead yields likelihoods over
			\(U_X\), which are used during the abduction step of counterfactual inference.
			
			Using the Gaussian quadrature rule
			\(
			\{u_s,\omega_s^{(U)}\}_{s=1}^{S},
			\)
			parent likelihood is approximated at the quadrature nodes
			\(
			\mathbf{x}_{\mathbf q}
			\)
			by
			\[
			\mathcal{B}_X[m_X](\mathbf{x}_{\mathbf q})
			\approx
			\sum_{s=1}^{S}
			\omega_s^{(U)}
			p_{U_X}(u_s)\,
			\widetilde m_X(\mathbf{x}_{\mathbf q},u_s).
			\]
			The propagated likelihood is combined with the stored prior parent state,
			\(
			\widetilde p_{\operatorname{Pa}(X)}(\mathbf{x})
			=
			p_{\operatorname{Pa}(X)}(\mathbf{x})
			\mathcal{B}_X[m_X](\mathbf{x}).
			\)
			When \(m_X\) represents the likelihood induced by the observations
			\(\mathcal O\), normalizing this quantity yields the posterior parent state
			\(
			p_{\operatorname{Pa}(X)\mid\mathcal O}(\mathbf{x}),
			\)
			represented in the corresponding tensor-product B-spline space using the same
			quadrature-based projection procedure as in the forward pass.
			
			\paragraph{Observation likelihoods.}
			
			Observed evidence is incorporated through local likelihoods. For an observation
			\(Y=y^\star\), we define
			\(
			L_Y(y)=L(y^\star\mid y),
			\)
			where \(L(y^\star\mid y)\) is a Gaussian kernel centered at \(y^\star\) providing a smooth 
			approximation of the Dirac measure corresponding to exact conditioning.
			If a clique contains multiple observations, their likelihoods are 
			multiplied. For
			\(
			\mathcal{O}_C
			=
			\{\,Y_r=y_r^\star: Y_r\in C\,\},
			\)
			the local likelihood is
			\(
			\Lambda_C(\mathbf{x}_C)
			=
			\prod\nolimits_{Y_r\in\mathcal{O}_C}
			L_{Y_r}(x_r).
			\)
			
			\paragraph{Junction-tree propagation.}
			
			The forward pass provides prior joint spline states \(p_C\) for the cliques of
			the constructed junction tree. Because these states are clique marginals,
			separator messages are represented as likelihood ratios rather than posterior
			densities. This avoids repeatedly incorporating the separator prior when neighboring clique beliefs are combined.
			For a directed edge \(C\rightarrow D\), define
			\[
			\Lambda_{C\setminus D}(\mathbf{x}_C)
			=
			\Lambda_C(\mathbf{x}_C)
			\prod\nolimits_{B\in\mathcal N(C)\setminus\{D\}}
			m_{B\rightarrow C}(\mathbf{x}_{S_{BC}}).
			\]
			The corresponding updated clique state is
			\[
			p_C^{\,C\setminus D}(\mathbf{x}_C)
			\propto
			p_C(\mathbf{x}_C)\Lambda_{C\setminus D}(\mathbf{x}_C).
			\]
			After marginalizing the updated and prior clique states onto
			\(S_{CD}\), the outgoing message is
			\[
			m_{C\rightarrow D}(\mathbf{x}_{S_{CD}})
			\propto
			\frac{
				p_{S_{CD}}^{\,C\setminus D}(\mathbf{x}_{S_{CD}})
			}{
				p_{S_{CD}}(\mathbf{x}_{S_{CD}})
			}.
			\]
			Thus, clique and separator densities are represented by
			spline coefficients, while likelihood messages are evaluated as ratios of
			these spline densities.
			After the collect and distribute passes, the posterior associated with clique
			\(C\) is
			\[
			p_{C\mid \mathcal{O}}(\mathbf{x}_C)
			=
			\frac{1}{Z_C}
			p_C(\mathbf{x}_C)
			\Lambda_C(\mathbf{x}_C) \!\!
			\prod_{D\in\mathcal N(C)} \!\!
			m_{D\rightarrow C}(\mathbf{x}_{S_{DC}}).
			\]
			To represent an updated density in the spline basis or perform normalization, the corresponding projection 
			is performed using tensor-product Gaussian quadrature.

			\paragraph{Counterfactuals. } 
			Counterfactual inference follows the abduct-intervene-infer procedure.
			During the abduction step, the local backward operator is applied while
			retaining the relevant exogenous variables in the propagated state, yielding
			their posterior distributions conditioned on the observations (\emph{abduct}). The intervention is then
			applied by replacing the corresponding structural mechanisms
			(\emph{intervene}). Finally, the modified SCM is evaluated using the abducted
			exogenous distribution (\emph{infer}). This preserves the shared exogenous realization between
			the factual and counterfactual worlds.

			\subsection{Complexity Analysis}
			
			Let \(N\) denote the number of endogenous variables, \(K\) the number of
			B-spline basis functions and \(Q\) the number of quadrature nodes per dimension,
			and \(S\) the number of quadrature nodes used for the exogenous variable. For a node with \(M\)
			parents, the tensor-product representation contains \(K^M\) coefficients and
			uses a \(Q^M\)-point quadrature grid. The implementation exploits
			tensor-product structure throughout via tensor contractions and
			Kronecker-structured operators, avoiding explicit tensor-product mass
			matrices.
			
			\paragraph{Forward pass.}
			
			For a node with \(M\) parents, the dominant cost is evaluating the structural
			mechanism and parent density over the \(SQ^M\) quadrature pairs and assembling
			the \(K\) child moments, requiring
			\(
			\mathcal O(SQ^M K)
			\)
			operations. For an SCM with \(N\) endogenous variables and maximum parent count
			\(M_{\max}\), the total forward-pass complexity is
			\[
			\mathcal O\!\left(NSQ^{M_{\max}}K\right).
			\]
			
			\paragraph{Backward pass.} 
			
			Let \(\mathcal C\) denote the clique set of the junction tree,
			\(C_{\max}\) its maximum clique size, 
			and \(\tau\) the treewidth of the triangulated SCM graph. The largest
			junction-tree clique then has size \(C_{\max}=\tau+1\). A clique update evaluates local
			likelihoods, separator messages, and, where required, the local backward
			operator over a tensor-product quadrature grid. Excluding the projection solve,
			the worst-case complexity is
			\[
			\mathcal O\!\left(|\mathcal C|\,SQ^{C_{\max}}K\right).
			\]
			Collect-distribute sends two directed messages along each
			junction-tree edge, making this bound tight up to a constant factor.
			
			\paragraph{Space complexity.}
			
			Let \(\mathcal J\) denote the maintained joint states and \(d_J\) its
			dimension. Storing the spline coefficients requires
			\(\mathcal O(\sum_{J\in\mathcal J} K^{d_J})\) memory, while the largest
			tensor-product quadrature grid requires
			\(\mathcal O(d_{\max}Q^{d_{\max}})\), where \(d_{\max}\) is the maximum state
			dimension. Excluding optional operator caching, the total memory requirement is
			\[
			\mathcal O\!\left(
			\sum\nolimits_{J\in\mathcal J} K^{d_J}
			+
			d_{\max}Q^{d_{\max}}
			\right).
			\]

			\begin{figure}[t]
				\centering
				\includegraphics[width=\columnwidth]{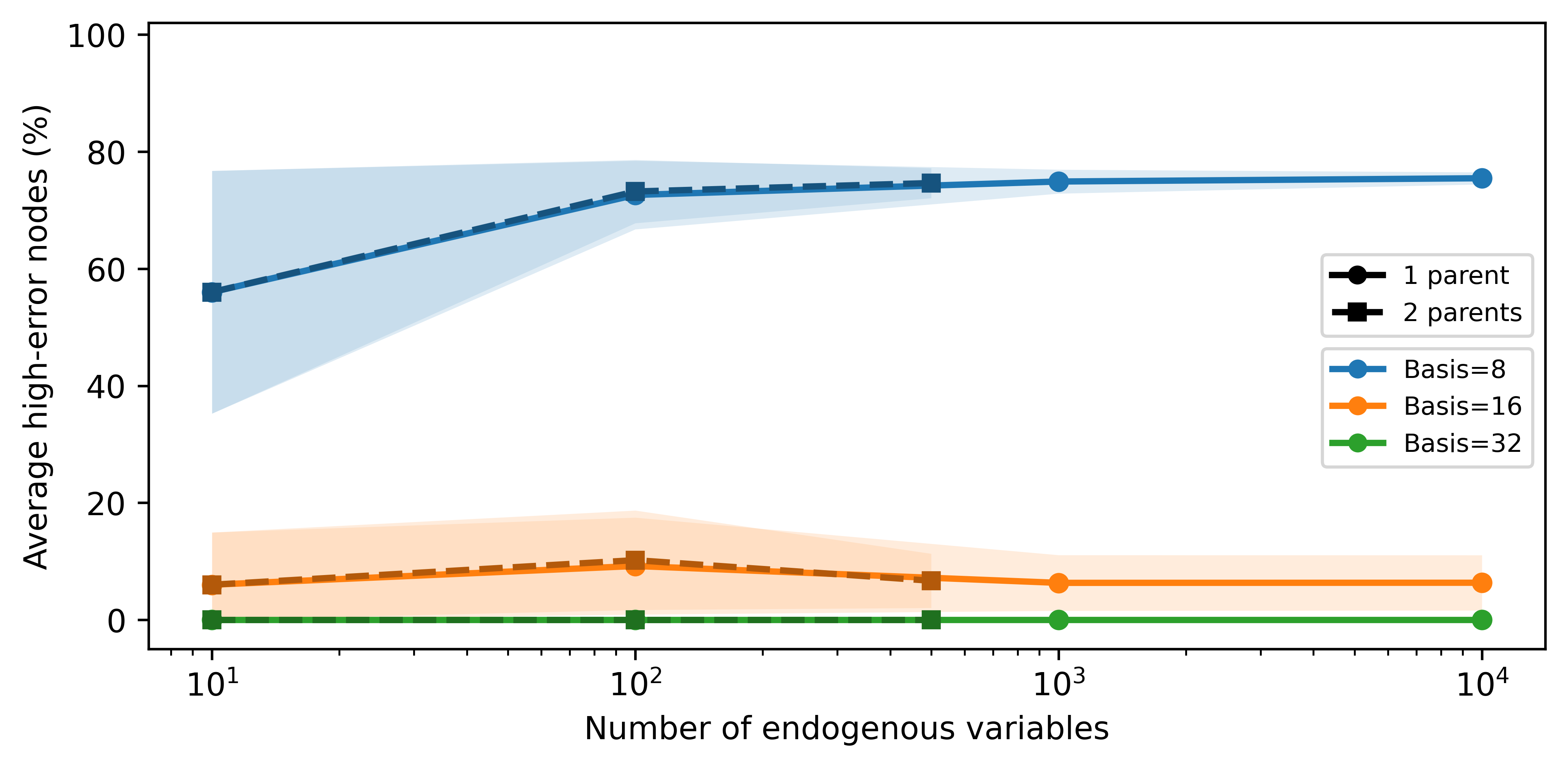}
				\caption{Mean percentage of endogenous variables with $\epsilon_{\mathrm{dom}} \geq 0.5\%$ in the forward pass versus the number of B-spline basis functions. Shaded bands show  standard deviation across independently generated tree and two-parent SCMs.}
				\label{fig:f_p1_err}
			\end{figure}

			\section{Experiments}
			
			For the experiments, we generate random directed acyclic SCMs with varying numbers of endogenous variables, 
			graph depths, and maximum parent orders. 
			For each configuration, we generate a random depth-constrained DAG. A
			single-parent backbone first fixes the graph depth, after which the remaining
			nodes are assigned random depths and connected to admissible parent sets from
			preceding levels, subject to the maximum in-degree.
			Exogenous noise distributions are sampled as random spline densities, 
			providing controlled variation in smoothness and multimodality. Endogenous densities and structural mechanisms 
			are represented using uniform cubic B-spline bases on $\Omega=[0,1]$, with Gauss-Legendre quadrature for numerical integration.
			
			The implementation exploits the resulting tensor-product structure through matrix-free tensor contractions and 
			Kronecker-structured operators. Quadrature-intensive computations are accelerated on 
			GPUs.\footnote{Implemented in PyTorch~2.11.0 with CUDA~12.8 on an NVIDIA RTX PRO 4000 Blackwell GPU; 8 CPU cores per task.} Parent-density, structural mechanism, basis, and projection evaluations are performed as batched tensor operations. Basis evaluations are cached and reused during inference.
			
			Inference is compared against an ancestral Monte Carlo (MC) reference computed from $10^6$
			independent samples. 
			Metrics are evaluated for every endogenous variable encountered during propagation and aggregated over the entire graph.
			For each endogenous variable we report two accuracy measures.
			For additional evaluation metrics, we refer the reader to the Appendix.
			The first is the domain-normalized expectation error,
			\[
			\epsilon_{\mathrm{dom}}
			=
			100\,
			\frac{
				\left|\mu_{\mathrm{ours}}-\mu_{\mathrm{MC}}\right|
			}{b-a},
			\]
			where $\Omega=[a,b]$ is the computational domain, measuring the expectation
			error relative to the variable's natural scale.
			The second is the Hellinger distance,
			\[
			H(p,q)
			=
			\frac{1}{\sqrt{2}}
			\left(
			\int_{\Omega}
			\left(
			\sqrt{p(x)}-\sqrt{q(x)}
			\right)^2
			\,dx
			\right)^{1/2},
			\]
			which compares the analytical density $p$ with the MC reference density $q$.
			It lies in $[0,1]$, where $0$ indicates identical densities.
			
			\begin{figure}[t]
				\centering
				\includegraphics[width=\columnwidth]{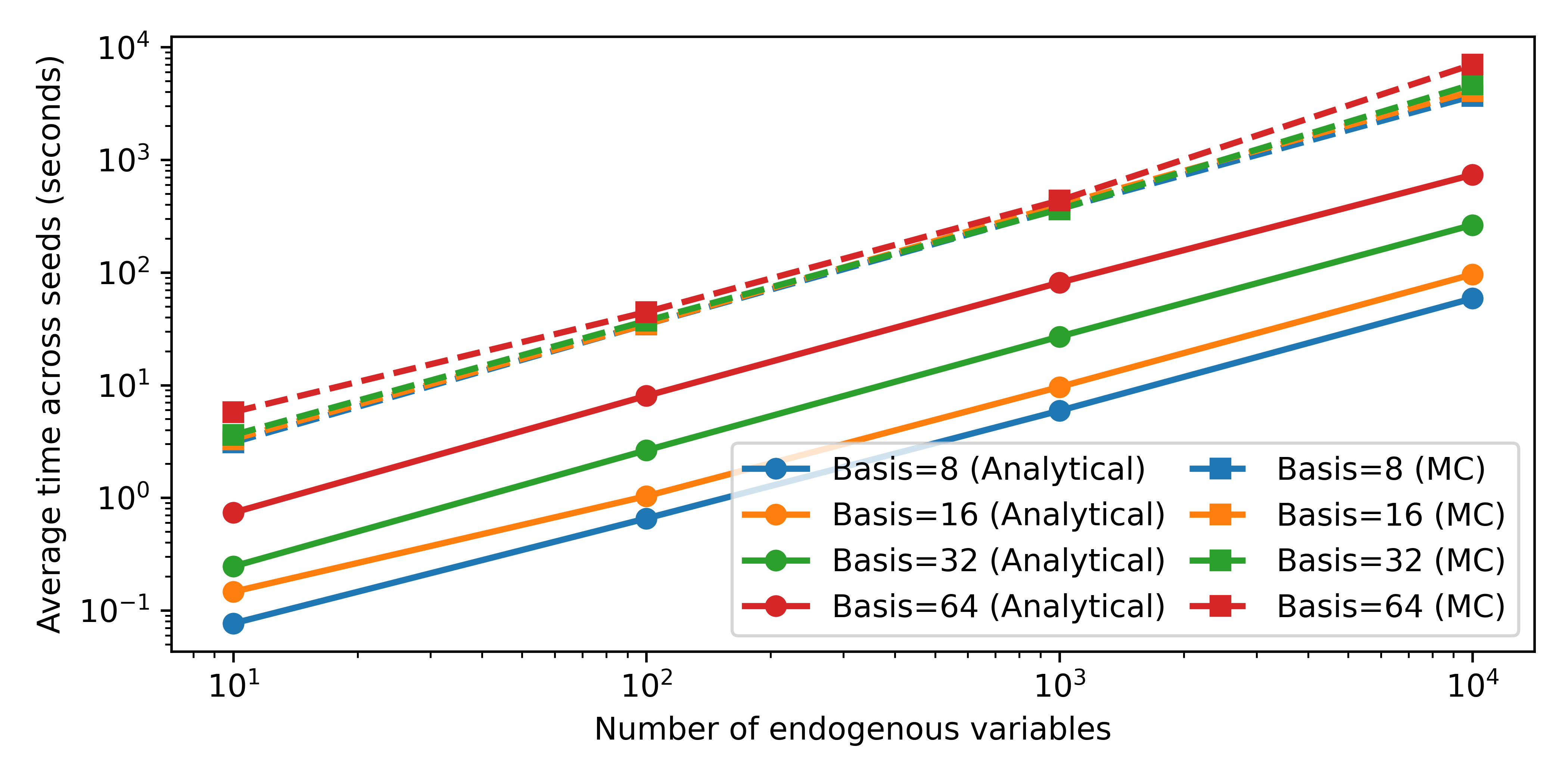}
				\caption{Average forward-pass runtime of the method and the MC reference versus the number of basis functions over randomly generated tree SCMs.}
				\label{fig:f_p1_time}
			\end{figure}
			
			We compare only against MC simulation since no existing method, to the best of
			our knowledge, performs general-purpose probabilistic inference in continuous
			SCMs directly on arbitrary nonlinear structural mechanisms. Existing approaches
			either target causal identification and estimation~\cite{dowhy:2020} or require
			compiling the SCM into an alternative tractable representation~\cite{causal_circuits:2022,wang:2023,ispn:2021}. 
			MC is therefore the natural model-agnostic baseline, converging to the true solution with increasing samples.

			\subsection{Results}
			
			The main experiments consider tree-structured SCMs and DAGs where approximately 20\% of the nodes have two parents,
			reflecting the exponential tensor-product scaling of the
			proposed representation. 
			We therefore focus on the regime where exact
			tensor-product inference remains computationally practical, leaving sparse and
			low-rank extensions to future work. Throughout, the number of quadrature points
			scales with the basis resolution; eight points per basis function were found
			best in preliminary experiments. For each graph size, we generate five
			random SCMs.

			\paragraph{Forward pass.}

\begin{figure}[t]
				\centering
				\includegraphics[width=\columnwidth]{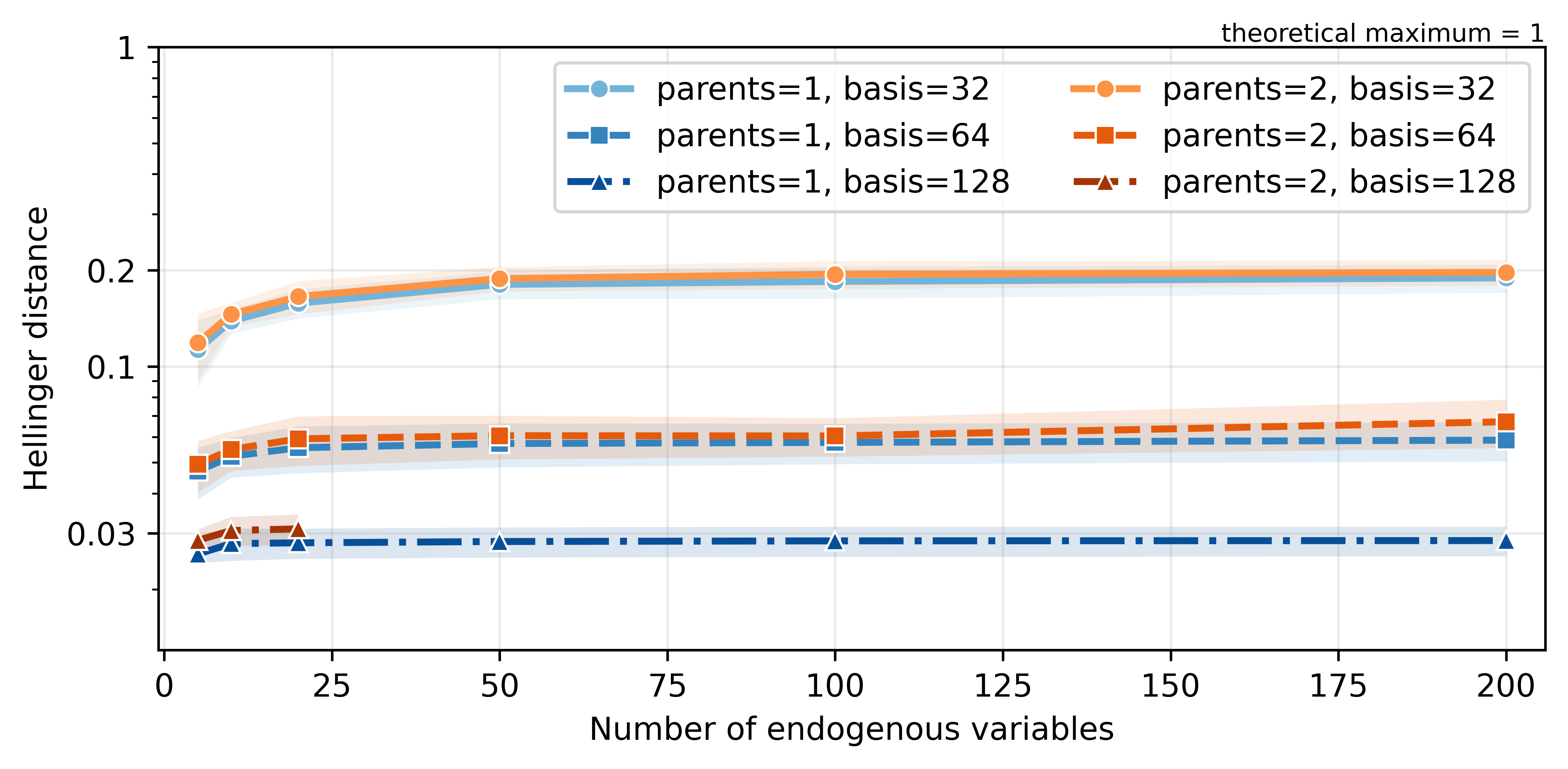}
				\caption{Hellinger distance between the estimated and reference forward densities.}
				\label{fig:dens_forw}
			\end{figure}

			Figure~\ref{fig:f_p1_err} reports the percentage of endogenous variables whose 
			domain-normalized expectation error exceeds $0.5\%$.  As expected, increasing 
			the number of basis functions consistently improves the approximation. 
			With $32$ basis functions, no high-error variables are observed, even for SCMs 
			with up to $10^4$ endogenous variables.
			For two-parent SCMs, computational cost limited experiments to graphs of at most $500$ endogenous variables, 
			and to only $20$ variables when using $128$ basis functions.
			For tree-structured SCMs, Figure~\ref{fig:f_p1_time} compares the runtime of the proposed 
			method with the high-accuracy MC reference. Our approach is approximately one 
			order of magnitude faster while maintaining comparable accuracy.
			Finally, Figure~\ref{fig:dens_forw} reports the mean Hellinger distance to the MC reference, 
			together with one standard deviation across experiments. The error decreases from approximately 
			$0.2$ with $32$ basis functions toward zero as the basis is refined, demonstrating convergence of the inferred densities.

			\begin{figure}[t]
				\centering
				\includegraphics[width=\columnwidth]{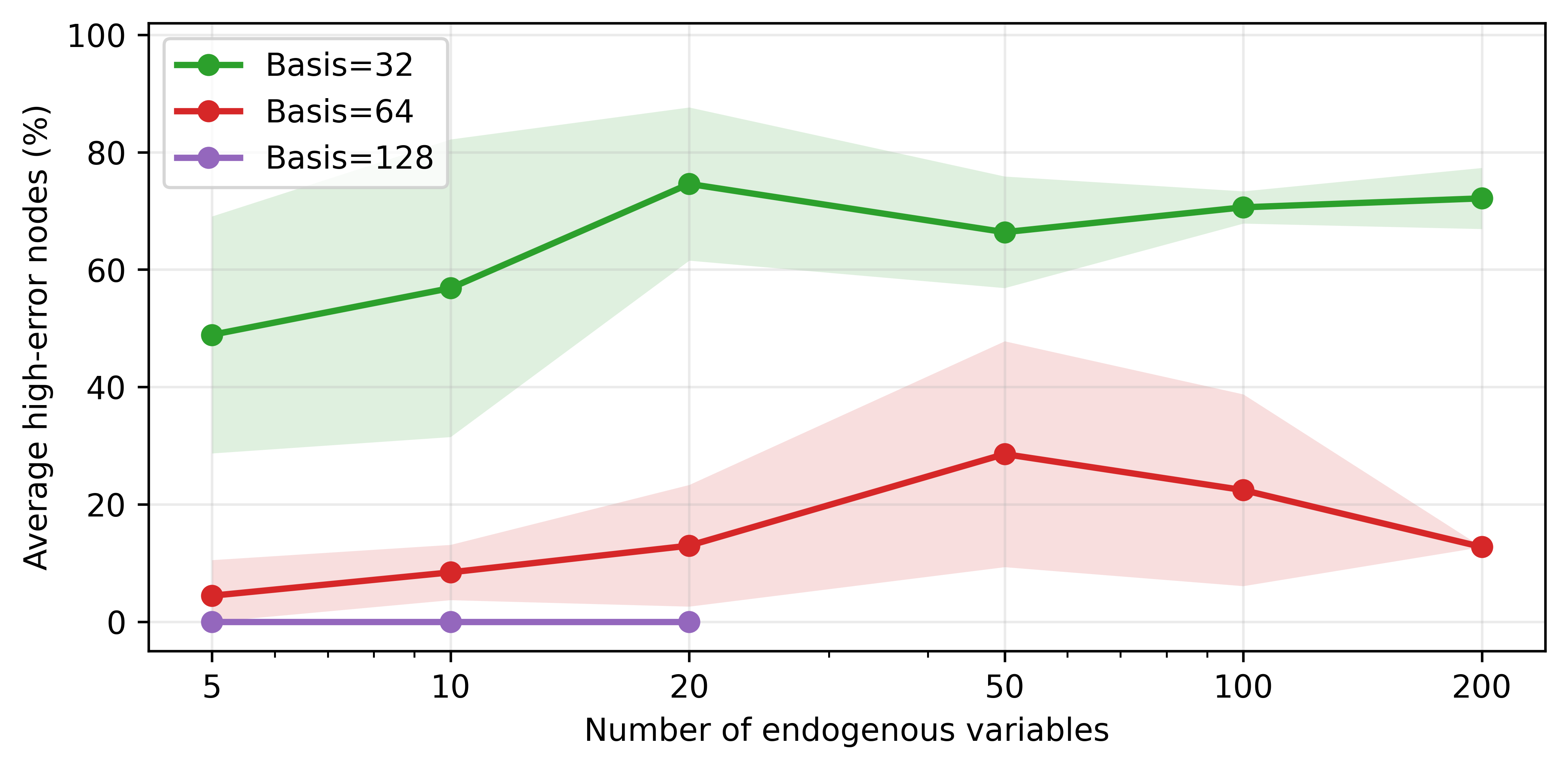}
				\caption{Mean percentage of endogenous variables with $\epsilon_{\mathrm{dom}} \geq 0.5\%$ in the backward pass for two-parent SCMs and 3 observed variables versus the number of basis functions.}
				\label{fig:b_p2_err}
			\end{figure}

			\begin{figure}[t]
				\centering
				\includegraphics[width=\columnwidth]{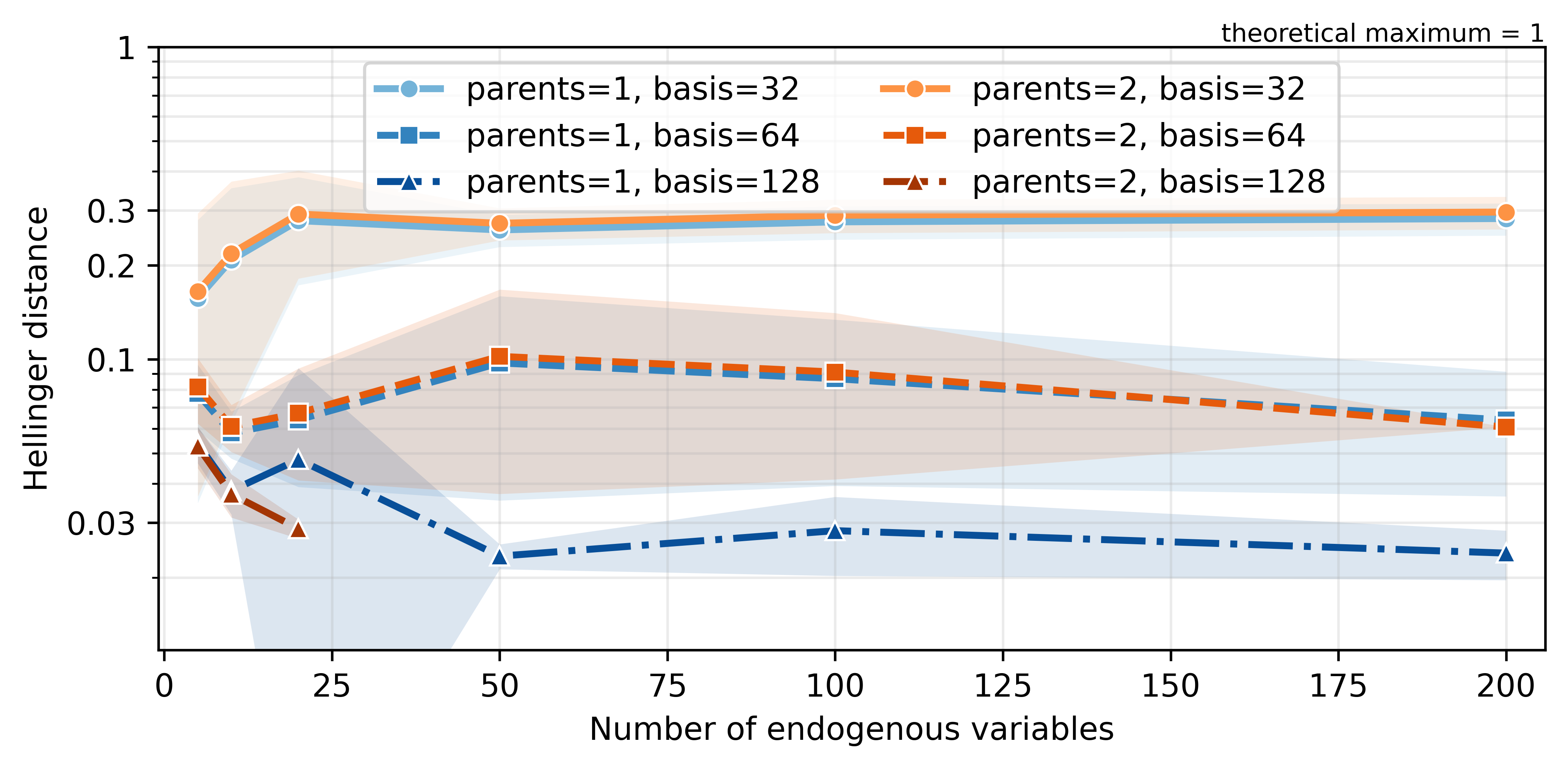}
				\caption{Hellinger distance between the estimated and reference backward densities for 3 observations. }
				\label{fig:dens_back:2}
			\end{figure}

			\paragraph{Backward pass.}
			
			For backward evaluation, we consider smaller SCMs than in the forward
			experiments, since conditioning evidence at leaf nodes becomes less informative
			for distant ancestors. Validating backward inference requires a different MC
			reference, as rejection sampling becomes impractical under conditioning.
			Instead, we generate ancestral samples by forward simulation and approximate
			conditioning using Gaussian kernel weights. Conditional expectations are
			estimated via self-normalized importance sampling.
			
			Figure~\ref{fig:b_p2_err} reports the average percentage of endogenous
			variables with \(\epsilon_{\mathrm{dom}}\ge0.5\%\) for two-parent SCMs with
			three observations, maximum clique size three, and maximum separator size two.
			Increasing the number of B-spline basis functions consistently improves the
			approximation, although backward inference requires a finer basis than the
			forward pass to eliminate high-error variables. This is expected, since
			conditional densities are more difficult to approximate and kernel weighting
			reduces the effective sample size of the MC reference. 
			Figure~\ref{fig:dens_back:2} reports the corresponding Hellinger distances for both tree and two-parent SCMs.
			Although backward inference exhibits larger errors and greater variability than
			the forward pass, the distance decreases systematically with basis resolution.
			As in the forward pass (Figure~\ref{fig:dens_forw}), the performance is largely
			independent of the number of parents.

			\section{Conclusion}
			
			This work introduced a message-passing framework for fully specified SCMs that
			performs observational, interventional, and counterfactual inference within a
			unified B-spline representation. By combining forward propagation of
			probability measures with backward propagation of likelihoods, the method avoids
			explicit construction of pushforward and conditional densities.
			Experiments demonstrate accurate probability densities and expectations 
			while scaling to large SCMs, with errors decreasing consistently as the spline resolution increases.
			Future work will investigate low-rank tensor
			approximations and adaptive basis refinement for higher-dimensional models with
			larger parent sets.

			\section*{Acknowledgments.}
			This research was funded in whole or in part by the Austrian Science Fund (FWF) (Standalone project "TraCause", 10.55776/PIN1217124). The work of Robert Peharz was additionally supported by the FWF Cluster of Excellence "Bilateral AI" (10.55776/COE12).
			
			\bibliography{aaai2027}
			
			
			\appendix
			
			\section{Causal Queries}
			
\begin{figure*}[t]
				\centering
				
				\tikzset{
					endo/.style={
						circle,
						draw,
						minimum size=5mm,
						inner sep=0pt,
						font=\scriptsize
					},
					exo/.style={
						rectangle,
						draw,
						minimum width=4mm,
						minimum height=2.8mm,
						inner sep=0.2pt,
						font=\tiny
					},
					intervened/.style={
						endo,
						fill=gray!30
					},
					causal/.style={
						-{Latex[length=1mm]},
						line width=0.35pt
					},
					modelbox/.style={
						draw,
						rounded corners=1.5pt,
						inner sep=1.5mm
					}
				}
				
				\begin{subfigure}[b]{0.23\linewidth}
					\centering
					\begin{tikzpicture}
						
						\node[endo] (S) {$S$};
						\node[endo, right=7mm of S] (T) {$T$};
						\node[endo, right=7mm of T] (C) {$C$};
						
						\node[exo, above=2.5mm of S] (US) {$U_S$};
						\node[exo, above=2.5mm of T] (UT) {$U_T$};
						\node[exo, above=2.5mm of C] (UC) {$U_C$};
						
						\draw[causal] (US) -- (S);
						\draw[causal] (UT) -- (T);
						\draw[causal] (UC) -- (C);
						
						\draw[causal] (S) -- (T);
						\draw[causal] (T) -- (C);
						
					\end{tikzpicture}
					\caption{Association}
					\label{fig:smoking-association}
				\end{subfigure}
				\hspace{0.15em}
				%
				\begin{subfigure}[b]{0.23\linewidth}
					\centering
					\begin{tikzpicture}
						
						\node[intervened] (S) {$S$};
						\node[endo, right=7mm of S] (T) {$T$};
						\node[endo, right=7mm of T] (C) {$C$};
						
						\node[exo, above=2.5mm of S] (US) {$U_S$};
						\node[exo, above=2.5mm of T] (UT) {$U_T$};
						\node[exo, above=2.5mm of C] (UC) {$U_C$};
						
						\node[font=\tiny, above left=-0.4mm and -0.2mm of S]
						{\(\mathit{const}\)};
						
						\draw[causal] (UT) -- (T);
						\draw[causal] (UC) -- (C);
						
						\draw[causal] (S) -- (T);
						\draw[causal] (T) -- (C);
						
					\end{tikzpicture}
					\caption{Intervention}
					\label{fig:smoking-intervention}
				\end{subfigure}
				\hspace{0.15em}
				%
				\begin{subfigure}[b]{0.50\linewidth}
					\centering
					\begin{tikzpicture}
						
						\node[endo] (S) {$S$};
						\node[endo, right=6mm of S] (T) {$T$};
						\node[endo, right=6mm of T] (C) {$C$};
						
						\draw[causal] (S) -- (T);
						\draw[causal] (T) -- (C);
						
						\node[modelbox, fit=(S)(T)(C)] (leftbox) {};
						
						\node[intervened, right=11mm of C] (Sp) {$S'$};
						\node[endo, right=6mm of Sp] (Tp) {$T'$};
						\node[endo, right=6mm of Tp] (Cp) {$C'$};

						\draw[causal] (Sp) -- (Tp);
						\draw[causal] (Tp) -- (Cp);
						
						\node[modelbox, fit=(Sp)(Tp)(Cp)] (rightbox) {};
						
						\coordinate (branchmid) at
						($(leftbox.north east)!0.5!(rightbox.north west)$);
						
						\node[exo] (US) at ($(branchmid)+(-5mm,10mm)$) {$U_S$};
						\node[exo, right=1mm of US] (UT) {$U_T$};
						\node[exo, right=1mm of UT] (UC) {$U_C$};
						
						\node[
						draw,
						rounded corners=1pt,
						fit=(US)(UC),
						inner sep=1.2mm
						] (Ubox) {};
						
						\draw[causal]
						($(Ubox.south west)!0.30!(Ubox.south east)$)
						--
						(leftbox.north);
						
						\draw[causal]
						($(Ubox.south west)!0.70!(Ubox.south east)$)
						--
						(rightbox.north);
						
					\end{tikzpicture}
					\caption{Counterfactual}
					\label{fig:smoking-counterfactual}
				\end{subfigure}
				
				\caption{
					Illustration of Pearl's causal hierarchy using the famous smoking-tar-lung
					cancer example-SCM, where \(S\) denotes smoking, \(T\) tar accumulation, and
					\(C\) lung-cancer risk.
					(a) Association concerns the observational distribution \(p(S,T,C)\).
					(b) Under \(do(S=s^\star)\), the mechanism of \(S\) is replaced by the
					constant assignment \(S:=s^\star\).
					(c) The counterfactual model shares the same exogenous realization across
					the factual and counterfactual branches.
				}
				\label{fig:smoking-causal-hierarchy}
			\end{figure*}
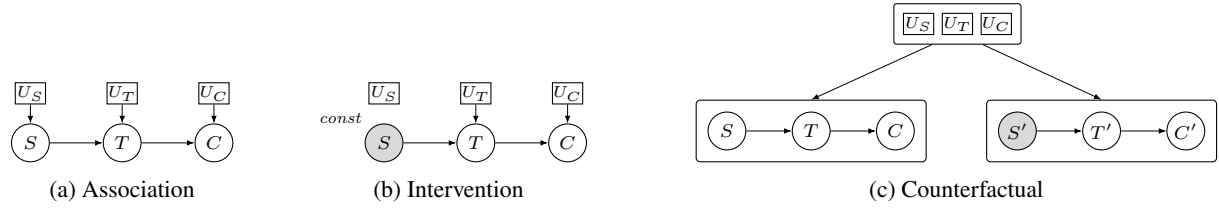

			Figure~\ref{fig:smoking-causal-hierarchy} illustrates the three levels of Pearl's causal hierarchy on a simple smoking-tar-lung cancer SCM. The observational model (Figure~\ref{fig:smoking-causal-hierarchy}a) represents the joint distribution induced by the structural mechanisms. An intervention (Figure~\ref{fig:smoking-causal-hierarchy}b) replaces the structural mechanism of the intervened variable by a constant assignment and removes its incoming causal edge. Counterfactual reasoning (Figure~\ref{fig:smoking-causal-hierarchy}c) is represented by a twin SCM, in which the factual and counterfactual worlds share the same exogenous variables while differing only in the applied intervention.
			
			\section{B-spline Basis Functions}\label{sec:bsplines}
			
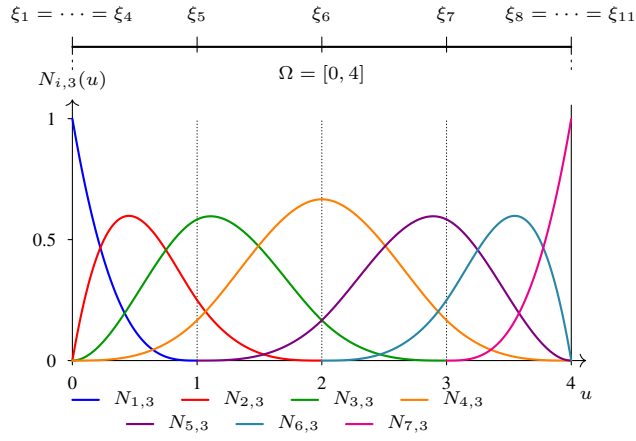
\begin{figure}[b]
				\centering
				\begin{tikzpicture}[
					x=1.65cm,
					y=3.2cm,
					line cap=round,
					line join=round,
					font=\scriptsize
					]
					
					\begin{scope}[yshift=4.65cm, y=1cm]
						
						\draw[thick] (0,-0.5) -- (4,-0.5);
						
						\foreach \x in {0,1,2,3,4}{
							\draw[thin] (\x,-0.57) -- (\x,-0.37);
						}
						
						\draw[thin] (4,-1.27) -- (4,-4.63);
						
						\foreach \y in {-0.60,-0.7,-0.80}{
							\draw[thin] (0,\y) -- (0,\y+0.025);
							\draw[thin] (4,\y) -- (4,\y+0.025);
						}
						
						\node[anchor=south] at (0,-0.3)
						{$\xi_1=\cdots=\xi_4$};
						
						\node[anchor=south] at (1,-0.3)
						{$\xi_5$};
						
						\node[anchor=south] at (2,-0.3)
						{$\xi_6$};
						
						\node[anchor=south] at (3,-0.3)
						{$\xi_7$};
						
						\node[anchor=south] at (4,-0.3)
						{$\xi_8=\cdots=\xi_{11}$};
						
						\node[anchor=north] at (2,-0.61)
						{$\Omega=[0,4]$};
						
					\end{scope}
					
					
					\draw[thin,->] (0,0) -- (4.12,0)
					node[below=8pt] {$u$};
					
					\draw[thin,->] (0,0) -- (0,1.08)
					node[above, anchor=south] {$N_{i,3}(u)$};
					
					\foreach \x in {0,1,2,3,4}{
						\draw[thin] (\x,0) -- (\x,-0.018);
						\node[anchor=north] at (\x,-0.025) {\x};
					}
					
					\foreach \y/\lab in {0/0,0.5/0.5,1/1}{
						\draw[thin] (0,\y) -- (-0.035,\y);
						\node[anchor=east] at (-0.055,\y) {\lab};
					}
					
					\foreach \x in {1,2,3}{
						\draw[black, densely dotted, very thin]
						(\x,0) -- (\x,1);
					}
					
					
					\draw[blue, thick]
					plot[domain=0:1, samples=80]
					(\x,{1 - 3*\x + 3*\x*\x - \x*\x*\x});
					
					\draw[red, thick]
					plot[domain=0:1, samples=80]
					(\x,{\x*(7*\x*\x - 18*\x + 12)/4});
					
					\draw[red, thick]
					plot[domain=1:2, samples=80]
					(\x,{-\x*\x*\x/4
						+ 3*\x*\x/2
						- 3*\x
						+ 2});
					
					\draw[green!60!black, thick]
					plot[domain=0:1, samples=80]
					(\x,{\x*\x*(18 - 11*\x)/12});
					
					\draw[green!60!black, thick]
					plot[domain=1:2, samples=80]
					(\x,{7*\x*\x*\x/12
						- 3*\x*\x
						+ 9*\x/2
						- 3/2});
					
					\draw[green!60!black, thick]
					plot[domain=2:3, samples=80]
					(\x,{-\x*\x*\x/6
						+ 3*\x*\x/2
						- 9*\x/2
						+ 9/2});
					
					\draw[orange, thick]
					plot[domain=0:1, samples=80]
					(\x,{\x*\x*\x/6});
					
					\draw[orange, thick]
					plot[domain=1:2, samples=80]
					(\x,{-\x*\x*\x/2
						+ 2*\x*\x
						- 2*\x
						+ 2/3});
					
					\draw[orange, thick]
					plot[domain=2:3, samples=80]
					(\x,{\x*\x*\x/2
						- 4*\x*\x
						+ 10*\x
						- 22/3});
					
					\draw[orange, thick]
					plot[domain=3:4, samples=80]
					(\x,{-\x*\x*\x/6
						+ 2*\x*\x
						- 8*\x
						+ 32/3});
					
					\draw[violet, thick]
					plot[domain=1:2, samples=80]
					(\x,{\x*\x*\x/6
						- \x*\x/2
						+ \x/2
						- 1/6});
					
					\draw[violet, thick]
					plot[domain=2:3, samples=80]
					(\x,{-7*\x*\x*\x/12
						+ 4*\x*\x
						- 17*\x/2
						+ 35/6});
					
					\draw[violet, thick]
					plot[domain=3:4, samples=80]
					(\x,{11*\x*\x*\x/12
						- 19*\x*\x/2
						+ 32*\x
						- 104/3});
					
					\draw[cyan!60!black, thick]
					plot[domain=2:3, samples=80]
					(\x,{\x*\x*\x/4
						- 3*\x*\x/2
						+ 3*\x
						- 2});
					
					\draw[cyan!60!black, thick]
					plot[domain=3:4, samples=80]
					(\x,{-7*\x*\x*\x/4
						+ 33*\x*\x/2
						- 51*\x
						+ 52});
					
					\draw[magenta, thick]
					plot[domain=3:4, samples=80]
					(\x,{\x*\x*\x
						- 9*\x*\x
						+ 27*\x
						- 27});
					
					\begin{scope}[yshift=-0.52cm, x=1cm, y=1cm]
						
						\draw[blue, thick] (0,0) -- (0.35,0);
						\node[anchor=west] at (0.42,0) {$N_{1,3}$};
						
						\draw[red, thick] (1.45,0) -- (1.80,0);
						\node[anchor=west] at (1.87,0) {$N_{2,3}$};
						
						\draw[green!60!black, thick] (2.90,0) -- (3.25,0);
						\node[anchor=west] at (3.32,0) {$N_{3,3}$};
						
						\draw[orange, thick] (4.35,0) -- (4.70,0);
						\node[anchor=west] at (4.77,0) {$N_{4,3}$};
						
						\draw[violet, thick] (0.72,-0.32) -- (1.07,-0.32);
						\node[anchor=west] at (1.14,-0.32) {$N_{5,3}$};
						
						\draw[cyan!60!black, thick] (2.17,-0.32) -- (2.52,-0.32);
						\node[anchor=west] at (2.59,-0.32) {$N_{6,3}$};
						
						\draw[magenta, thick] (3.62,-0.32) -- (3.97,-0.32);
						\node[anchor=west] at (4.04,-0.32) {$N_{7,3}$};
						
					\end{scope}
					
				\end{tikzpicture}
				
				\caption{Open cubic B-spline basis for
					\(\boldsymbol{\xi}=(0,0,0,0,1,2,3,4,4,4,4)\)
					on \(\Omega=[0,4]\).}
				\label{fig:bspline_basis}
			\end{figure}

			The proposed framework employs tensor-product B-spline basis functions to
			represent probability densities over bounded computational domains.
			Throughout this work, let
			\[
			\Omega=[a,b]\subset\mathbb{R}
			\]
			denote the computational domain of a single variable. A univariate B-spline
			basis is defined by a non-decreasing knot vector
			\[
			\boldsymbol{\xi}
			=
			(\xi_1,\ldots,\xi_m),
			\qquad
			\xi_i\le\xi_{i+1},
			\]
			and polynomial degree \(p\). If the spline space contains \(K\) basis
			functions, then
			\(
			m=K+p.
			\)
			Throughout this work, we employ open knot vectors, in which the first and last
			knots are repeated \(p+1\) times. Consequently, the spline basis is defined on
			\[
			\Omega=[\xi_{p+1},\xi_{m-p}],
			\]
			which coincides with the interval bounded by the first and last distinct knots.
			The B-spline basis
			\[
			\{N_{i,p}\}_{i=1}^{K}
			\]
			is defined recursively by the Cox-de Boor formula. For
			\(p=0\),
			\[
			N_{i,0}(u)
			=
			\begin{cases}
				1, & \xi_i\le u<\xi_{i+1},\\[1mm]
				0, & \text{otherwise},
			\end{cases}
			\]
			\begin{align*} N_{i,p}(u) &= \frac{u-\xi_i}{\xi_{i+p}-\xi_i} N_{i,p-1}(u)\\ &+ \frac{\xi_{i+p+1}-u} {\xi_{i+p+1}-\xi_{i+1}} N_{i+1,p-1}(u), \end{align*}
			where fractions with zero denominator are interpreted as zero. Figure~\ref{fig:bspline_basis}
			illustrates an open cubic B-spline basis.
			Any spline in the resulting approximation space can be expressed as
			\[
			s(u)
			=
			\sum_{i=1}^{K}
			w_iN_{i,p}(u),
			\]
			where \(w_i\in\mathbb{R}\) are the spline coefficients.
			
			For multivariate functions, the computational domain is
			\[
			\Omega=\Omega_1\times\cdots\times\Omega_d
			\subset\mathbb{R}^d.
			\]

			\section{Isogeometric Analysis}\label{sec:isogeometric}
			
			Isogeometric analysis (IGA) employs spline basis functions for numerical approximation. The proposed framework adopts several numerical techniques from IGA, including spline representations and efficient quadrature for evaluating integrals.
			For this section, without loss of generality, we omit the degree notation in the B-spline basis.
			
			We first consider the one-dimensional case. Let
			$f \colon\Omega\rightarrow\mathbb{R}$ be approximated in a finite-dimensional spline space spanned by the B-spline basis
			\(
			\{N_i\}_{i=1}^{n}.
			\)
			The approximation is
			\[
			\hat f(x)
			=
			\sum\nolimits_{i=1}^{n}
			w_iN_i(x)
			=
			\mathbf N(x)^\top\mathbf w,
			\]
			where
			\(
			\mathbf N(\mathbf{x})
			=
			[N_1(\mathbf{x})\ \cdots\ N_n(\mathbf{x})]^\top
			\)
			denotes the basis vector and
			\(
			\mathbf w=(w_1,\ldots,w_n)^\top
			\)
			contains the spline coefficients.
			The coefficients are determined through the $L^2(\Omega)$ projection by minimizing
			\[
			J(\mathbf w)
			=
			\frac12
			\int_\Omega
			\left(
			f(\mathbf{x})
			-
			\mathbf N(\mathbf{x})^\top\mathbf w
			\right)^2
			\,d\mathbf{x}.
			\]
			The first-order optimality conditions yield the linear system
			\[
			\mathbf M\mathbf w
			=
			\mathbf b,
			\]
			where
			\(
			M_{ij}
			=
			\int_\Omega
			N_i(\mathbf{x})
			N_j(\mathbf{x})
			\,d\mathbf{x}
			\)
			are the entries of the \emph{mass matrix} and
			\(
			b_i
			=
			\int_\Omega
			N_i(\mathbf{x})f(\mathbf{x})
			\,d\mathbf{x}
			\)
			the \emph{load vector} ones. 
			
			The multivariate case is obtained by tensor products of the one-dimensional bases. Let $N_{\mathbf i}$ be a multivariate spline.
			Then the entries of the mass matrix become
			\[
			M_{\mathbf i\mathbf j}
			=  
			\prod\nolimits_{k=1}^{d}
			\int_\Omega
			N^{(k)}_{i_k}(x_k)
			N^{(k)}_{j_k}(x_k)\,dx_k,
			\]
			where the equality follows from the tensor-product structure of the basis. Consequently, multidimensional quadrature reduces to tensor products of one-dimensional Gaussian quadrature rules. These rules are exact whenever the integrand is a polynomial and the quadrature order is chosen to integrate its degree exactly.
			
			\section{Additional Evaluation}
			
\begin{figure}[b]
				\centering
				\includegraphics[width=\columnwidth]{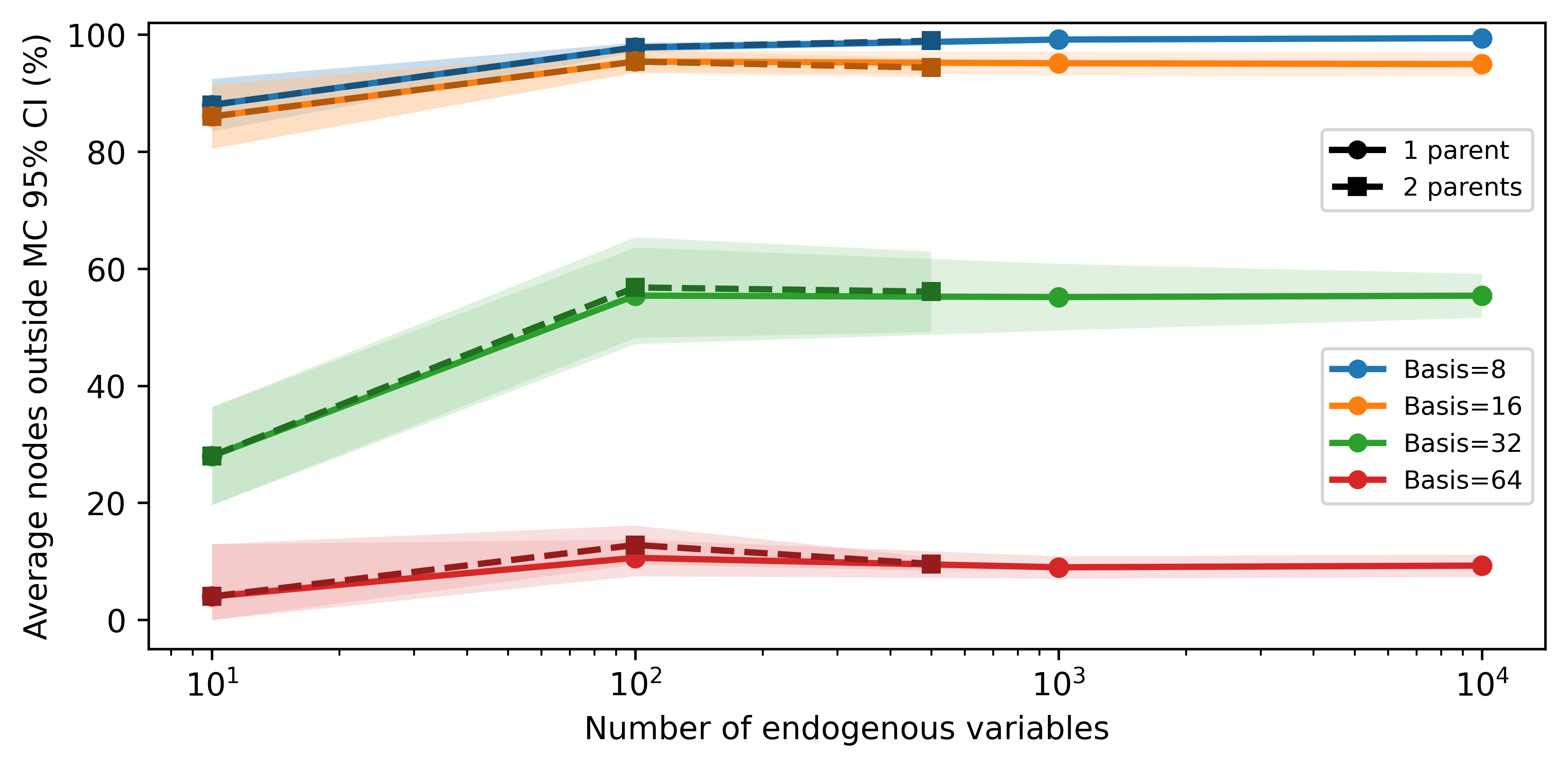}
				\caption{Mean percentage of endogenous variables whose analytical expectation lies within the $95\%$ MC confidence interval in the forward pass, as a function of the number of B-spline basis functions.}
				\label{fig:f_p1_mc}
			\end{figure}

			Along with the domain-normalized error, for the forward pass we record whether the analytical expectation lies within the
			$95\%$ confidence interval of the MC estimator, 
			assessing whether any discrepancy is consistent with the sampling
			uncertainty of the MC reference.
			Figure~\ref{fig:f_p1_mc} reports the percentage of endogenous variables whose 
			analytical expectation falls outside the MC $95\%$ confidence interval. Across 
			experiments, agreement with the MC reference 
			remains stable as graph size increases and improves monotonically with the basis resolution.
			
		\end{document}